\documentclass[runningheads]{llncs}

\usepackage{graphicx}

\usepackage{graphicx}
\usepackage{subcaption}
\usepackage{hyperref}       
\usepackage{url}            
\usepackage{booktabs}       
\usepackage{amsfonts}       
\usepackage{nicefrac}       
\usepackage{microtype}      
\usepackage{color}
\usepackage{float}
\usepackage{authblk}
\usepackage{amsmath}
\usepackage{hyperref} 
\usepackage{fullpage}
\usepackage{lipsum}
\usepackage{float}
\usepackage{xcolor}
\usepackage{bm}
\usepackage{bbold}
\usepackage{amssymb}
\usepackage{comment}

\usepackage{bibunits}

\makeatletter
\renewcommand{\@biblabel}[1]{[#1]}  
\makeatother

\renewcommand{\thetable}{\arabic{table}}

\begin{document}

\begin{bibunit}

\title{False Promises in Medical Imaging AI? \\ Assessing Validity of Outperformance Claims}

\titlerunning{False Promises in Medical Imaging AI?}

\author{Evangelia Christodoulou\inst{1,2} \and Annika Reinke\inst{1,2,3} \and Pascaline Andrè \inst{4} \and Patrick Godau\inst{1,2,5,18} \and Piotr Kalinowski\inst{1,2,5,18} \and Rola Houhou\inst{1,2}  \and Selen Erkan\inst{6} \and Carole H. Sudre\inst{7,8} \and Ninon Burgos\inst{4} \and Sofiène Boutaj\inst{4,9} \and Sophie Loizillon\inst{4} \and Maëlys Solal\inst{4}  \and Veronika Cheplygina\inst{10} \and Charles Heitz \inst{4} \and Michal Kozubek \inst{11} \and Michela Antonelli\inst{8,12} \and Nicola Rieke\inst{13} \and Antoine Gilson \inst{4} \and Leon D. Mayer\inst{1,2}  \and Minu D. Tizabi\inst{1,2}  \and M. Jorge Cardoso\inst{12} \and Amber Simpson\inst{14,15}  \and Annette Kopp-Schneider\inst{16}\and Gaël Varoquaux\inst{17,\star} \and Olivier Colliot\inst{4,\star} \and Lena Maier-Hein\inst{1,2,3,18,19,\star}}

\institute{German Cancer Research Center (DKFZ) Heidelberg, Div. Intelligent Medical Systems, Germany 
  \email{evangelia.christodoulou@dkfz-heidelberg.de} \and 
 National Center for Tumor Diseases (NCT), NCT Heidelberg, a partnership between DKFZ and Heidelberg University Hospital, Germany
 \and DKFZ Heidelberg, Helmholtz Imaging, Germany 
 \and Sorbonne Université, Institut du Cerveau - Paris Brain Institute - ICM, CNRS, Inria, Inserm, AP-HP, Hôpital de la Pitié-Salpêtrière, France 
 \and
  HIDSS4Health - Helmholtz Information and Data Science School for Health, Germany
 \and DKFZ Heidelberg, Interactive Machine Learning Group, Germany
  \and MRC Unit for Lifelong Health and Ageing at UCL and Centre for Medical Image Computing, Department of Computer Science, University College London, UK
   \and School of Biomedical Engineering and Imaging Science, King’s College London, UK
 \and MICS, CentraleSupelec, Paris-Saclay University, France
   \and Department of Computer Science, IT University of Copenhagen, Denmark
    \and Centre for Biomedical Image Analysis, Masaryk University, Brno, Czech Republic
 \and  Centre for Medical Image Computing, University College London, UK
  \and NVIDIA, Germany
 \and School of Computing, Queen’s University, Canada
 \and Department of Biomedical and Molecular Sciences, Queen’s University, Canada
  \and Division of Biostatistics, DKFZ,  Germany
 \and Parietal project team, INRIA Saclay-Île de France, France 
 \and Faculty of Mathematics and Computer Science, Heidelberg University, Germany
  \and Medical Faculty, Heidelberg University, Germany }

 \renewcommand{\thefootnote}{\fnsymbol{footnote}}
 \footnotetext[1]{\tiny Shared last authors: L. Maier-Hein, O. Colliot, and G. Varoquaux}

  \authorrunning{E. Christodoulou et al.}

\maketitle

\section*{Abstract}

Performance comparisons are fundamental in medical imaging Artificial Intelligence (AI) research, often driving claims of superiority based on relative improvements in common performance metrics. However, such claims frequently rely solely on empirical mean performance. In this paper, we investigate whether newly proposed methods genuinely outperform the state of the art by analyzing a representative cohort of medical imaging papers. We quantify the probability of false claims based on a Bayesian approach that leverages reported results alongside empirically estimated model congruence to estimate whether the relative ranking of methods is likely to have occurred by chance. According to our results, the majority ($>$80\%) of papers claims outperformance when introducing a new method. Our analysis further revealed a high probability ($>$5\%) of false outperformance claims in 86\% of classification papers and 53\% of segmentation papers. These findings highlight a critical flaw in current benchmarking practices: claims of outperformance in medical imaging AI are frequently unsubstantiated, posing a risk of misdirecting future research efforts.

\section*{Main}

The field of medical imaging Artificial Intelligence (AI) has experienced exponential growth, with thousands of papers published annually today. Despite this surge in research, clinical practice has not witnessed a corresponding transformation, as most proposed AI models remain unadopted in real-world healthcare settings.

\begin{figure}[H]
\centering
\includegraphics[width=0.8\textwidth]{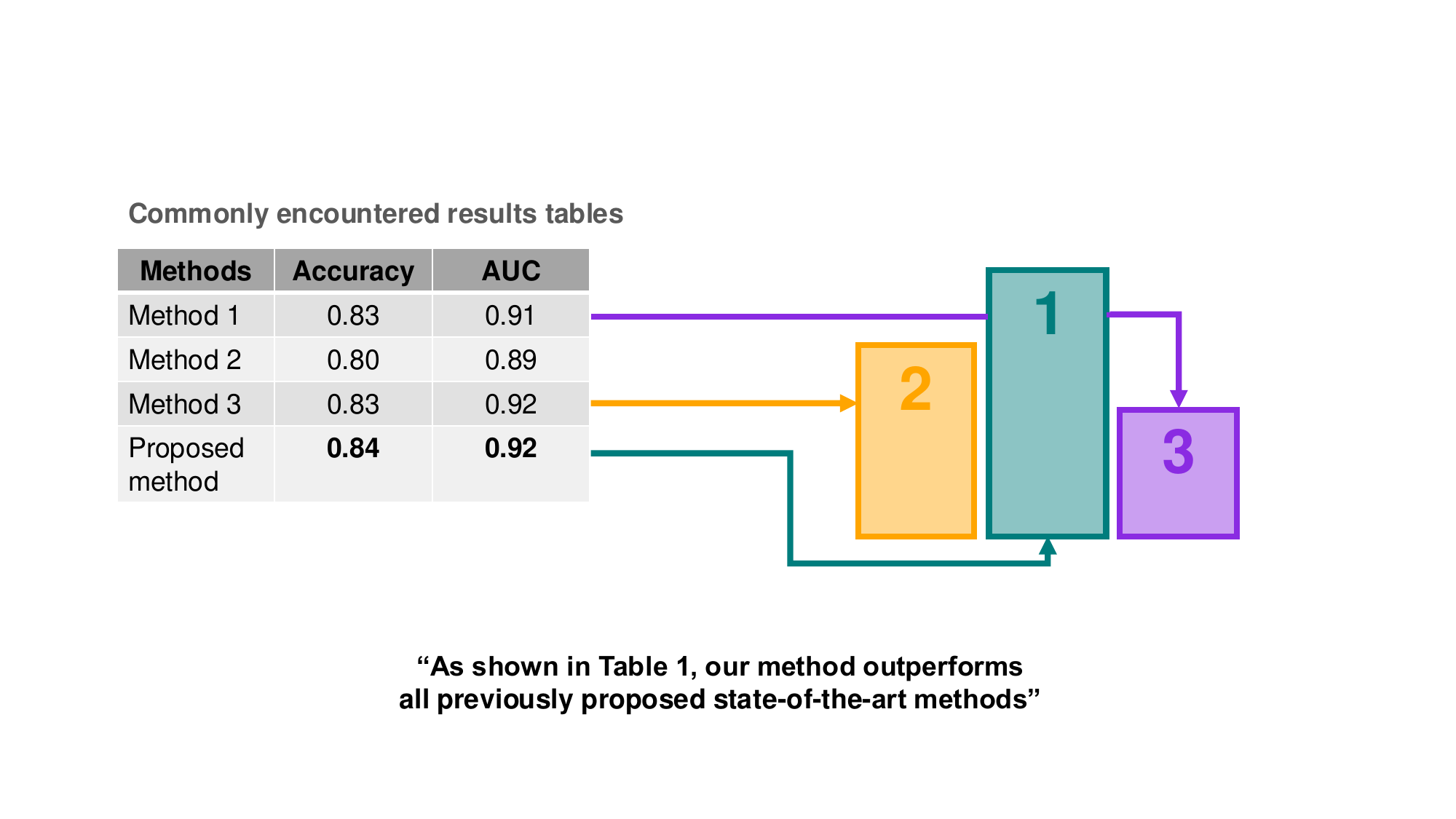} 
\caption{\textbf{\textit{Example of commonly encountered results tables in biomedical imaging analysis publications.}} Claims of superiority are often based on mean performance metric estimates alone, indicated by bold-face numbers.\\
AUC; Area Under the Curve}
\label{Fig1}
\end{figure}

In this paper, we hypothesize that misleading conclusions drawn from performance evaluations contribute to this gap. Medical imaging AI studies frequently present performance comparisons to claim superiority over competing methods. However, these claims are often based on empirical mean performance alone, without accounting for variability due to data fluctuations (see Figure \ref{Fig1}). A widespread reporting practice involves highlighting the highest mean performance metric values in bold, without displaying measures of uncertainty such as standard deviations or confidence intervals. This issue is not unique to medical imaging AI but is also prevalent in the broader AI field, where research, including AI methods for medical imaging, is most commonly published in top-tier machine learning conferences such as the Conference on Neural Information Processing Systems (NeurIPS), the International Conference on Learning Representations (ICLR), and the IEEE/CVF Conference on Computer Vision and Pattern Recognition (CVPR).

The core question we address in this paper is: Are common claims of outperformance in medical imaging AI well-substantiated? Or is there a high risk of false promises being propagated through published research?

\begin{figure}[H]
\centering
\includegraphics[width=0.8\textwidth]{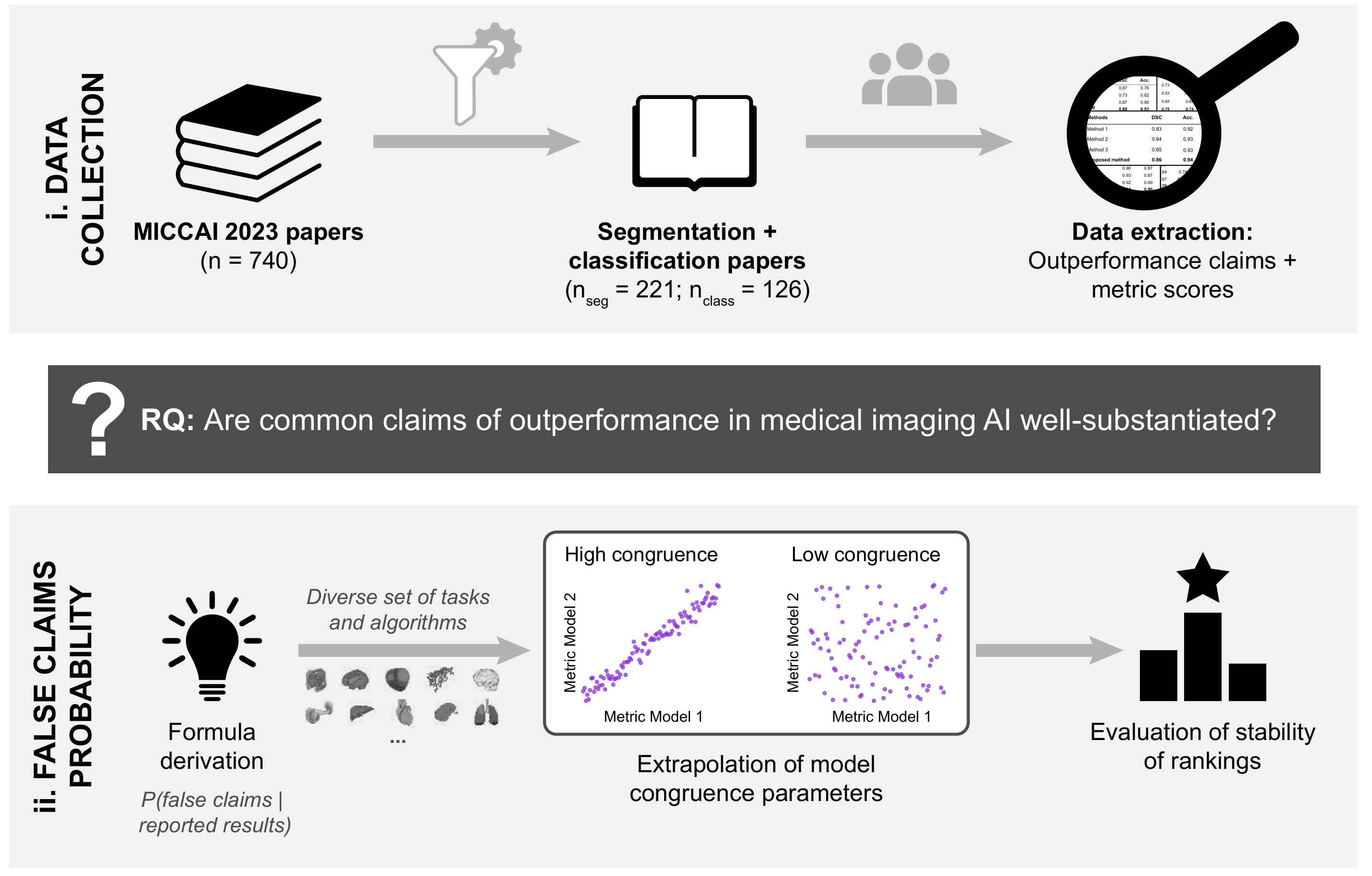} 
\caption{\textbf{\textit{Methods used to investigate the prevalence of unsubstantiated outperformance claims in medical imaging Artificial Intelligence (AI).}} Top: Based on all MICCAI 2023 classification and segmentation publications, we systematically analyzed the evidence based on which outperformance claims are made in medical image classification and segmentation. Bottom: Parameters reflecting model congruence were estimated based on a diverse set of tasks and algorithms. Based on these, we employed a Bayesian approach to compute the probability of false claims.\\
MICCAI; Medical Image Computing and Computer-Assisted Intervention conference
}
\label{Fig2}
\end{figure}

Our specific contribution is twofold (Figure \ref{Fig2}): Firstly, using a representative cohort of 347 medical imaging AI publications, we systematically analyze the evidence based on which outperformance claims are made. Secondly, to investigate whether a method truly outperforms the state of the art, we compute the probability of false claims—an estimate of whether observed rankings are likely to have occurred by chance—using a Bayesian approach. With these contributions, we are the first to reveal the high prevalence of unsubstantiated superiority claims in medical imaging AI, thereby highlighting critical flaws in current benchmarking practices.

\section*{Results}

\subsection*{Medical image machine learning (ML) research abundantly features papers including results tables with bold-face numbers}

To investigate common reporting practices in medical imaging AI research, we conducted a systematic analysis of performance reporting across a representative cohort of publications from the Medical Image Computing and Computer-Assisted Intervention (MICCAI) 2023 conference. MICCAI is an ideal venue to assess benchmarking standards in medical imaging AI as it is recognized for its highly competitive review process and emphasis on methodological advancements over state-of-the-art solutions. In 2023, MICCAI received over 2,300 submissions, accepting 730 papers (30\% increase in submissions compared to 2022). In 2024, submissions rose to 2,869, with 857 papers provisionally accepted, maintaining an acceptance rate of approximately 30\%. These figures underscore the conference's selectivity and its role in showcasing high-quality research in the field.

\begin{figure}[H]
\centering
\includegraphics[width=0.8\textwidth]{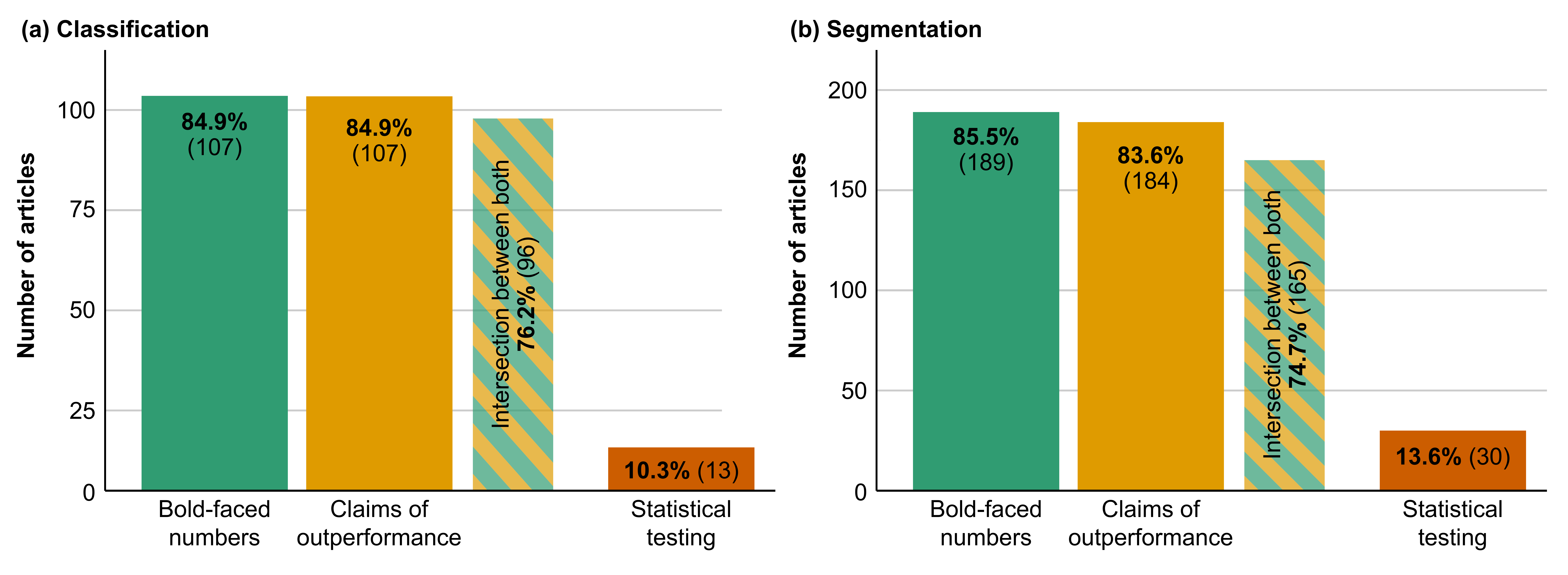} 
\caption{\textbf{\textit{Claims of outperformance in medical classification (a) and segmentation (b) papers are not well-justified.}} The majority of papers ($>$ 80\%) claims outperformance and highlights the best performance metric values in bold. A low portion ($<$15\%)  backs up the claims by further analyses, such as statistical testing.
}
\label{Fig3}
\end{figure}

Our analysis revealed that over 80\% of MICCAI 2023 classification and segmentation papers highlight the results of the top-performing method in bold-faced numbers, using this formatting as visual reinforcement for claims of methodological superiority (Figure \ref{Fig3}). However, despite these strong claims of outperformance, only 13\% of classification papers and 10\% of segmentation papers attempted to justify their conclusions by performing statistical significance testing.

When it came to performance differences, the median Accuracy value was 0.01 and the median Dice Similarity Coefficient (DSC) value was 0.01 for classification and segmentation papers respectively (Figure S3). Among the papers that employed a test set in their data-splitting strategy (65\% of classification papers and 76\% of segmentation papers), the median test set sizes were 500 images (Q1–Q3: 98 – 4970) for classification papers and 62 images (Q1–Q3: 25 – 223) for segmentation papers. 

\subsection*{Common sample sizes lead to high probability of false claims}

To evaluate whether reported AI superiority over state-of-the-art methods is quantitatively reliable, we applied a Bayesian approach to estimate the likelihood that observed performance rankings could have occurred by chance (see Methods). This method quantifies uncertainty in benchmarking to identify unreliable outperformance claims. To this end, we focused on the most widely applied metrics---Accuracy and DSC, which were by far the most widely used metrics for classification (72\%; see Table S1) and segmentation (95\%; see Table S2) respectively. Whether or not a given performance delta is actually meaningful depends not only on its magnitude but also on how well model outputs are aligned. A key factor influencing false claims is therefore a quantity we call model congruence (see Supplementary Information, Subsection D)---a measure of performance alignment amongst methods. For segmentation tasks, congruence is represented by the correlation of metric values achieved by different models (here: DSC), while for classification tasks it is defined as the proportion of cases where both models concurrently made correct predictions.
 
 As we did not have access to the raw experimental data of the publications analyzed and congruence values are typically not reported in papers, we took an empirical approach to derive reasonable estimates of congruence values. Using a diverse set of medical image segmentation and classification tasks spanning a wide range of different medical imaging modalities as well as a representative set of models (see Methods), we derived descriptive statistics for the congruence values and used the median for the following analyses.
 
\begin{figure}[H]
\centering
\includegraphics[width=0.8\textwidth]{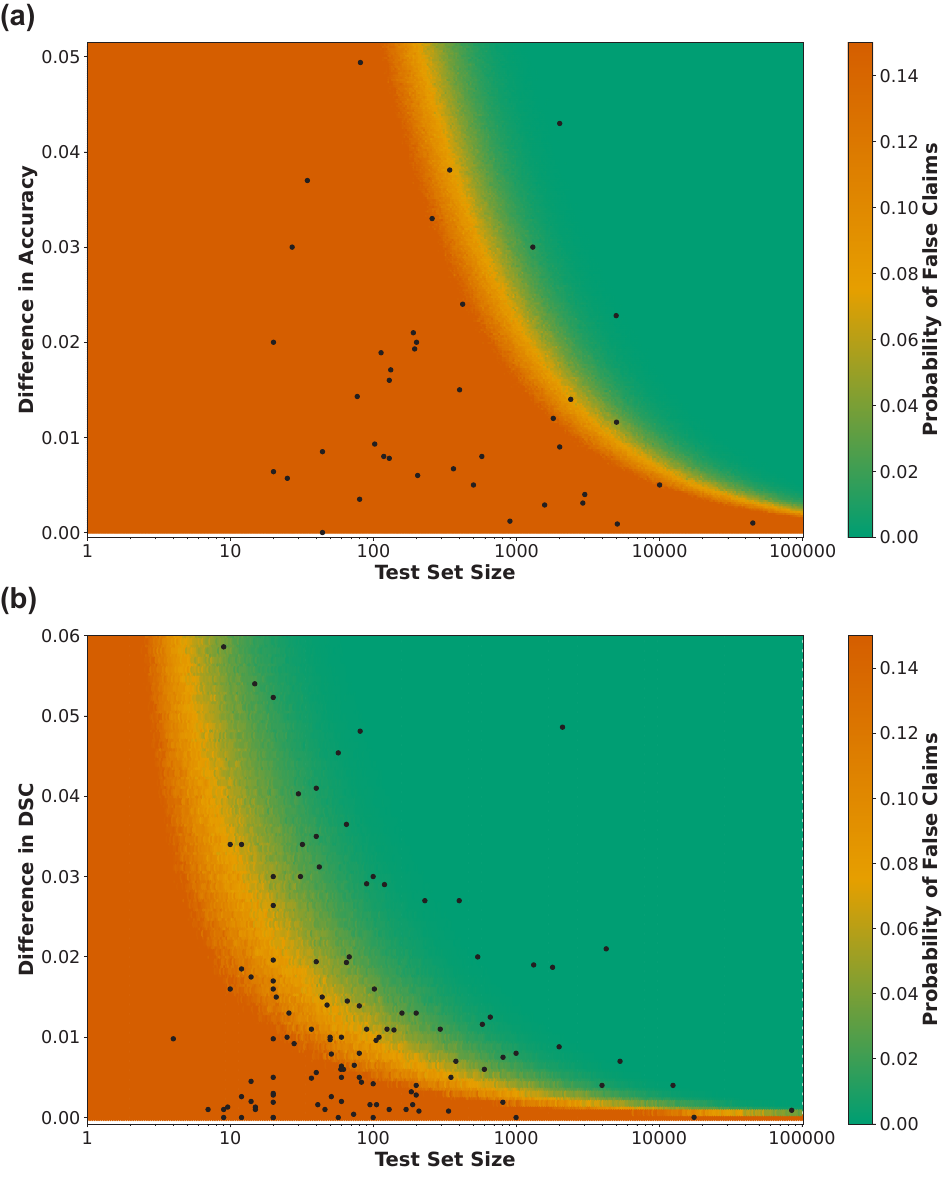} 
\caption{\textbf{\textit{Our analysis reveals a high likelihood for false outperformance claims in medical image classification and segmentation.}} (a) For more than 86\% of all MICCAI 2023 classification papers, we estimated a high probability ($>$5\%) of false claims of outperformance of their first-ranked method over the second-ranked one. The dots correspond to the reported difference in Accuracy between the two top-ranked methods of the MICCAI 2023 classification papers and their respective test set sizes.  (b) For more than 53\% of all MICCAI 2023 segmentation papers, we estimated a high probability ($>$5\%) of false claims of outperformance of their first-ranked method over the second-ranked one. The dots correspond to the reported difference in DSC between the two top-ranked methods of the MICCAI 2023 segmentation papers and their respective test set sizes. For both figures, the x-axis, representing the test set size, is displayed on a logarithmic scale.}
\label{Fig4}
\end{figure}

Our analysis revealed a high probability ($>$5\%) of false claims in the majority of medical imaging AI publications. Specifically, 86\% of classification papers (Figure \ref{Fig4}a) and 53\% of segmentation papers (Figure \ref{Fig4}b) exhibited a false claim probability exceeding this threshold, when computing the results based on the median congruence values. Further, for 58\% of classification papers (Figure \ref{Fig5}a) and 25\% of segmentation papers (Figure \ref{Fig5}b) we even found an extremely high probability of false claims of over 30\%.

\newpage

\subsection*{Our conclusions are robust with respect to the estimation of  assumed congruence parameters}

\begin{figure}[H]
\centering
\includegraphics[width=0.8\textwidth]{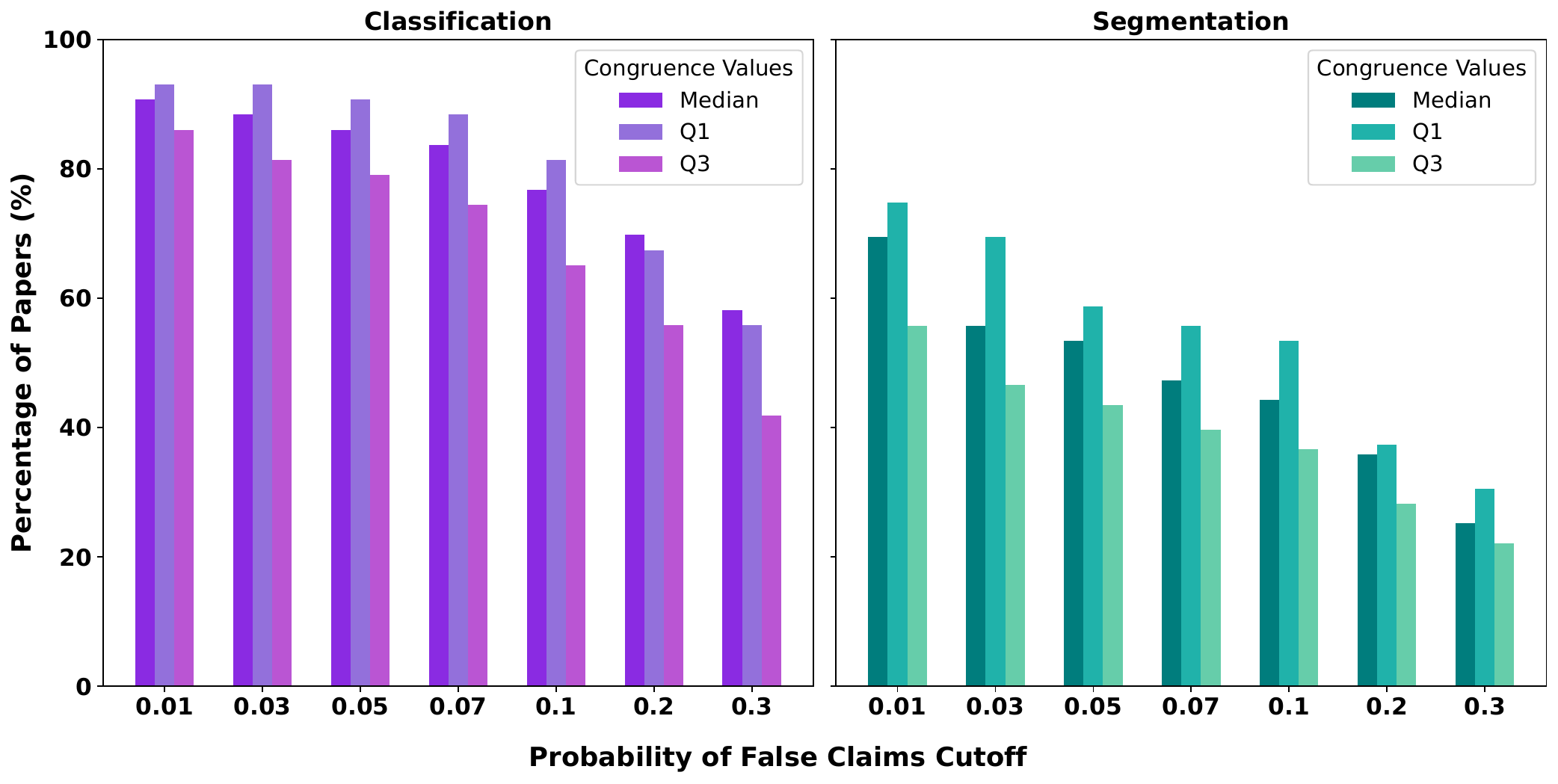} 
\caption{\textbf{\textit{The high likelihood of false outperformance claims in medical image classification and segmentation remains robust across a range of assumed congruence parameters and false claim probability thresholds.}} Cumulative percentage of MICCAI 2023 classification papers and segmentation papers  that are exceeding various cutoffs of the probability of false claims for a variety of  congruence parameter values (median, Q1 or Q3). 
\\Q1; 25th percentile, Q3; 75th percentile, MICCAI; Medical Image Computing and Computer-Assisted Intervention conference
}
\label{Fig5}
\end{figure}

To assess the robustness of our conclusions---that the majority of MICCAI 2023 classification and segmentation papers have a high probability of outperformance claims, we repeated our analysis with the first quartile (Q1) and third quartile (Q3) of the original congruence values obtained from these datasets.  Our results, summarized in Figure \ref{Fig5}, demonstrate that our conclusions remain robust regardless of the specific choice of assumed congruence parameters and the cutoffs used for the probability of false claims.

\subsection*{Stronger evidence of outperformance calls for test sets dramatically larger than usual}

\begin{figure}[H]
\centering
\includegraphics[width=0.8\textwidth]{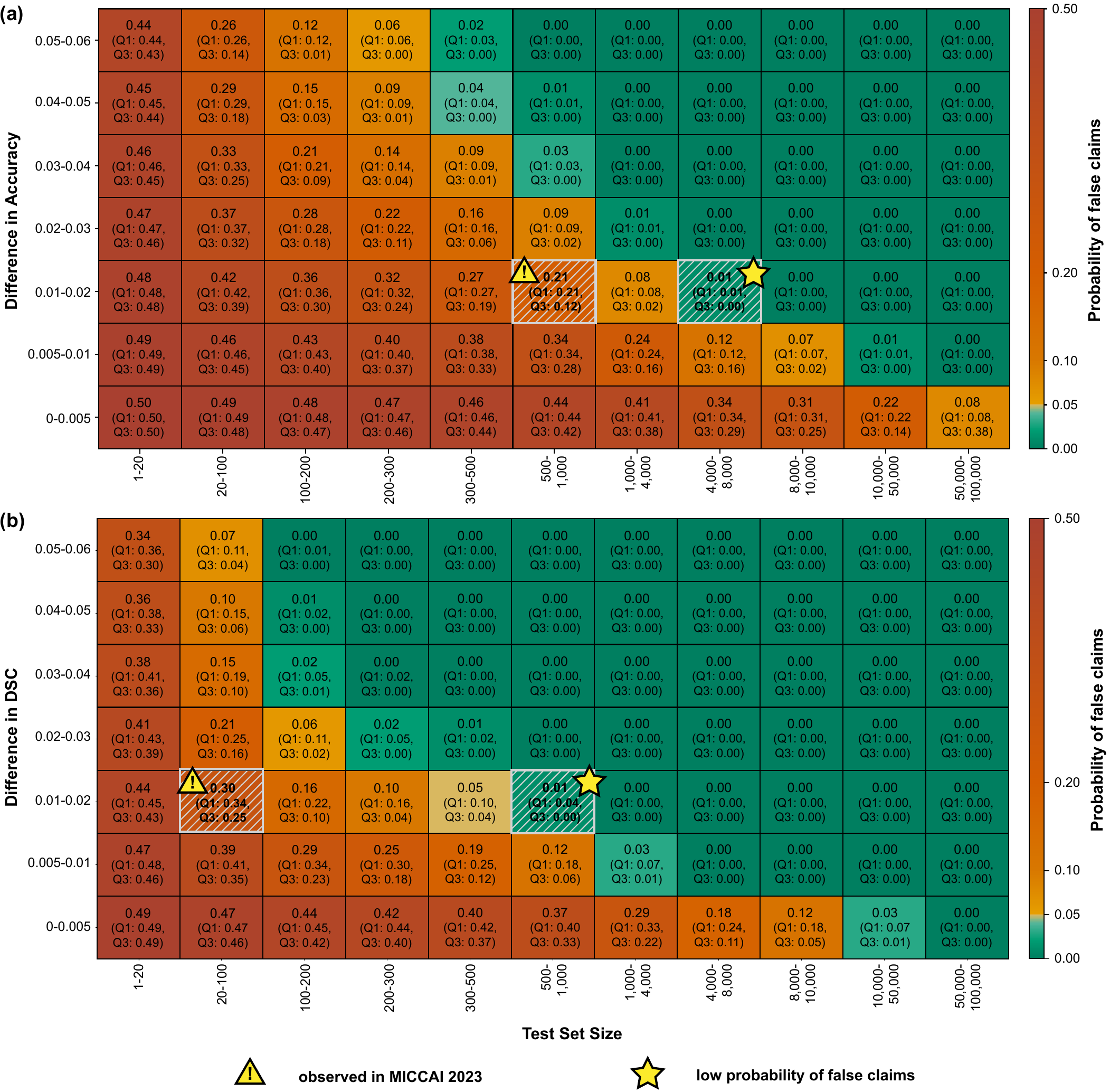} 
\caption{\textbf{\textit{Substantiating outperformance requires substantially larger sample sizes.}} Probability of false claims as a function of test set size and metric difference for (a) classification papers and (b) segmentation papers. Each cell displays the median probability based on assumed median model congruence, with values for the 25th and 75th percentile model congruence in parentheses. Color coding: red ($>$10\%), orange (5–10\%), green ($<$5\%). The left gray-highlighted cell represents observed MICCAI 2023 values, while the right gray-highlighted cell indicates test set sizes leading to $<$5\% false claim probability for a fixed range of performance deltas. 
\\Q1; 25th percentile, Q3; 75th percentile, MICCAI; Medical Image Computing and Computer-Assisted Intervention conference
}
\label{Fig6}
\end{figure}

Our analysis reveals that typical median performance differences (deltas) in segmentation and classification papers are around 0.01 (Figure S3). This value is likely to arise by chance given common sample sizes (see Figure \ref{Fig4}). To assess how many test samples are needed to substantiate claims of outperformance at this level, we conducted simulation studies across various performance deltas, sample sizes, and typical model congruence values. Results show that demonstrating a 0.01 Accuracy improvement requires a test set at least 8× larger than the median 500 images (see Figure \ref{Fig6}a), while a 10× larger test set is needed for a 0.01 DSC difference in segmentation (see Figure \ref{Fig6}b). This pattern holds across both tasks, underscoring the need for significantly larger evaluation datasets to ensure credible performance claims in medical imaging AI research.

\newpage

\section*{Discussion}

This study is the first to systematically quantify the likelihood of false outperformance claims in medical imaging AI research, revealing the high prevalence of unsubstantiated superiority claims. Our findings clearly suggest that common sample sizes used in medical imaging AI studies often lack the statistical power needed to support strong conclusions of superiority in performance. The failure to account for uncertainty and variability in performance evaluation raises concerns about the credibility of benchmarking practices, highlighting the urgent need for more rigorous statistical assessment and transparent reporting in AI-driven medical imaging research.

Our framework also offers practical support for reviewers in assessing whether the articles they are evaluating may include unsubstantiated claims of outperformance. For example, reviewers can refer to Figures~\ref{Fig5} and \ref{Fig6} as initial lookup tables to quickly estimate the probability of false claims based on reported test set sizes and performance differences. For a more comprehensive assessment, they can compute actual probabilities based on raw experimental data using the formulas provided in the Methods section (see Supplementary Information, Subsection C).

Our method, which estimates the probability of false outperformance claims using a Bayesian framework, is conceptually related to traditional statistical significance testing but differs in several key aspects. Both approaches aim to assess how likely it is that an observed performance difference has occurred by chance; however, their underlying principles and interpretations diverge significantly. A key distinction is that traditional statistical significance testing operates within a frequentist framework. The corresponding p-value is the probability that the difference in performance between methods is at least as strong as the one reported, if the performance of the proposed method was in fact not better than the state-of-the art (i.e., if the null hypothesis (H0) was true). It should not be interpreted as the probability that the null hypothesis is true. On the contrary, our Bayesian approach estimates the probability that the proposed method is not better than the state of the art, given the results reported in the paper. Under a non-informative prior, the Bayesian formula is similar to the frequentist one but the interpretation differs.

To ease interpretation of our results, we reported the percentage of papers for which the probability of false claims is higher than 5\%, corresponding to the most widely used threshold in statistics. Nevertheless, the percentage of papers with potential false claims is alarmingly high across a wide range of thresholds and thus our conclusions would hold under different thresholds. The fact that the probability of false claims exceeds 0.3 for large portions of 58\% (classification) and 25\% (segmentation) of the papers is extremely alarming, suggesting that many newly suggested methods do not provide a benefit compared to previously proposed methods.

False outperformance claims in medical imaging AI research often stem from factors that give models an unfair advantage during evaluation, making them seem more effective than they actually are. Two major contributing factors for this are model overfitting and---more generally---data leakage, both of which lead to overly optimistic performance estimates and poor generalization to new data. Model overfitting often occurs when authors repeatedly test different models on identical test sets and then report on the performance of the best model. Data leakage is a more general concept that refers to a situation where information that should not be available during the model training phase is inadvertently included, leading to overly optimistic performance estimates or inaccurate predictions. This occurs because the model learns patterns or relationships from the “leaked” data. Common types of leakage include target leakage, train-test-contamination (e.g., when preprocessing steps like normalization or feature engineering are performed on the entire dataset before splitting it into training and test sets) and temporal leakage in time-series data \cite{kapoor2023leakage}. Small sample sizes make these problems worse as they undermine statistical power of model evaluation in AI \cite{varoquaux2018cross}. Publication bias, with corresponding incentives and selective reporting, are known to make statistical problems worse \cite{ioannidis2005most}. Our results reveal the pervasiveness of these failures in medical image AI \cite{varoquaux2018cross}. 

The likelihood of false outperformance claims was higher for classification than for segmentation, despite classification studies generally having larger test sample sizes (median = 500 images) compared to segmentation studies (median = 62 images) (Figure S2). While this may initially seem counterintuitive, it is a natural consequence of how performance information is derived in each task. In classification, each image annotation contributes only a single bit of information regarding performance, whereas in segmentation, every voxel or pixel provides one bit, which is then aggregated at the image level. This results in segmentation tasks effectively utilizing a much larger amount of information per image, reducing the likelihood of false claims despite smaller test set sizes.

A limitation of our study could be seen in the fact that we did not have access to the true congruence parameters. To estimate the probability of false claims, we relied on external datasets to approximate these key parameters for both segmentation and classification. For segmentation, we used the Medical Segmentation Decathlon (MSD) dataset to calculate the correlation coefficient of the test set DSC and to approximate unreported standard deviation values in the analyzed MICCAI 2023 papers \cite{antonelli2022medical}. For classification, we extrapolated information on the assumed congruence from a comprehensive private dataset. While these choices were driven by pragmatic considerations, and generalizability could be a potential limitation, our sensitivity analyses confirm that the methodology remains robust across varying parameter assumptions.

In conclusion, our findings emphasize the urgent need for more rigorous benchmarking practices in medical imaging AI research. The prevalent reliance on mean performance metrics, without accounting for variability or conducting statistical significance testing, leads to a high probability of false claims of model superiority. This not only misguides the research community but also potentially hampers the clinical adoption of AI models by creating misleading expectations of performance improvements. To address this issue, future studies must incorporate robust statistical analyses, report performance variability, and utilize sufficiently large test sets to substantiate claims of outperformance. Without these improvements, the field risks perpetuating unreliable conclusions, ultimately slowing progress toward trustworthy and clinically impactful AI solutions.

\section*{Methods}

\subsection*{Data collection}

We obtained data through a systematic analysis of classification and segmentation papers published at the MICCAI 2023 conference. The initial set of papers was identified by searching for the terms “classification” or “segmentation” in the title or abstract of all MICCAI 2023 publications. As a general inclusion criterion, we focused on papers where classification or segmentation constituted the main task; papers that did not meet this criterion were considered out of scope. From the initial set of 133 classification papers and 252 segmentation papers, we excluded 7 classification papers and 31 segmentation papers for being out of scope for our analysis, resulting in 126 classification papers and 221 segmentation papers from MICCAI 2023 (Figure \ref{Fig2}). For each eligible paper, we extracted information on claims of outperformance, data splitting strategies, reported performance metrics, and whether performance uncertainty was addressed (e.g., through reporting standard deviations or confidence intervals). A team of 12 researchers conducted data extraction for classification papers, and 13 researchers for segmentation papers. Data extraction was performed independently by pairs of researchers. In cases where discrepancies arose, conflicts were resolved through discussion with two additional researchers (EC and AR) who were not involved in the initial data extraction process.

\subsection*{Data analysis subset}
For the general performance variability analysis, we focused on the set of papers that satisfied the general inclusion criteria described above. Prior to conducting our quantitative analysis, we excluded papers that reported performance metric values only in figures or supplementary materials. This exclusion affected 7\% of classification papers and 5\% of segmentation papers. For the calculation of the probabilities of false claims, we applied additional selection criteria. Specifically, we included only papers that (1) incorporated a data splitting strategy involving a distinct test set---this included splits denoted as “validation” or “test” in a single train/test or train/validation setup,  three-way splits consisting of train, validation, and test sets, or test sets coming in addition to a k-fold cross-validation (therefore we had to exclude papers which only only cross-validation without independent test); (2) reported Accuracy values for at least two methods in classification tasks or DSC values for at least two methods in segmentation tasks; and (3) provided the test set size for each method considered within the paper. These criteria resulted in a final dataset of 43 classification papers and 131 segmentation papers from MICCAI 2023 for our probability of false claims analysis.

\subsection*{Probability of false claims formulas}

To estimate whether a method truly yielded a benefit over the state of the art, we relied upon a Bayesian approach to estimate whether the relative ranking of methods is likely to have occurred by chance. This approach resulted in the formulas explained below, with detailed derivations presented in Supplementary Information (Subsection C).

To estimate whether a method truly yielded a benefit over the state-of-the-art, we relied upon a Bayesian approach to estimate whether the relative ranking of methods is likely to have occurred by chance. This approach resulted in the formulas explained in the following, with detailed derivations presented in the Supplementary Information.

For a given paper,  the "probability of false claims" is defined as the probability that the second-ranked method was, in fact, performing equally or better than the first-ranked method, given the results reported in the paper. This can be formulated in a Bayesian setting as: 
$$
P(\theta_A \leq \theta_B | \textbf{reported results})=P(\theta_A \leq \theta_B | \widehat{\theta}_A, \widehat{\theta}_B)$$

where $\theta_A$ (resp. $\theta_B$) is the true performance of the first-ranked method A (resp. second-ranked method B), treated as a random variable, and $\widehat{\theta_A}$ (resp. $\widehat{\theta_B}$) is the performance of method A (resp. method B) reported in the paper. 
Note that, by definition, the probability of false claims cannot be greater than $0.5$, since one has observed that $\widehat{\theta_A}>\widehat{\theta_B}$.

In the following, we provide two formulas for the probability of false claims: 
\begin{itemize}
    \item for classification tasks: the performance metric we focus on is Accuracy, i.e., the probability that a sample is well classified, and we classically write $\theta= p$.
    \item for segmentation tasks: the performance is the mean DSC and we classically write $\theta=\mu$.
\end{itemize} 

\subsubsection*{Formula for classification}

The probability of false claims for classification is defined as: 
\begin{align}
 P(\theta_A \leq \theta_B \mid \textbf{reported results})=P(p_A \leq p_B \mid  \widehat{p}_A, \widehat{p}_B)=P(p_1 \leq p_2 \mid \widehat{p}_A, \widehat{p}_B) = \int_{0}^{1} \int_{0}^{p_2}p(p_1, p_2| \widehat{p}_A, \widehat{p}_B) dp_1 dp_2
\label{eqn:integral}
\end{align}
where :
\begin{itemize}
    \item $p_A$ (resp. $p_B$) is the probability that a given set is correctly classified by method A (resp. method B).
    \item $p_1$ (resp. $p_2$) is the probability that a given set is well classified by method A (resp. method B) and not by method B (resp. method A).
    \item $p(p_1, p_2| \widehat{p}_A, \widehat{p}_B)= D\left(x_1 + 1, x_2 + 1, n - x_1 - x_2 + 2\right)$, where $D$ is the Dirichlet distribution. This distribution naturally arises as a conjugate prior of the multinomial distribution which models the likelihood of the different proportions, such as $p_1$ and $p_2$.
    \item $n$ is the sample size of the test set.
    \item $\widehat{p}_A$ (resp. $\widehat{p}_B$ ) the reported accuracy of method A (resp. method B).
    \item $\widehat{p}_{1,1}$ is the model congruence, i.e., the proportion of cases where both methods made correct predictions. As we do not have access to this value, we make an assumption based on previous experiments.
    \item $x_1=n(\widehat{p}_A-\widehat{p}_{1,1})$ 
    \item $x_2=n(\widehat{p}_B-\widehat{p}_{1,1})$ 
\end{itemize}
The integral~\ref{eqn:integral} is not tractable analytically and was therefore computed using Monte Carlo sampling. Specifically, one samples $k$ times $p_1,p_2$ from the Dirichlet distribution and counts the number of times $M$ where $p_1\leq p_2$. Then the probability of false claims is approximated by: 
$$P(p_A \leq p_B \mid  \widehat{p}_A, \widehat{p}_B) \approx \frac{M}{k}$$

\subsubsection*{Formula for segmentation}
    
\label{subsec:defDSC}
The probability of false claims for segmentation is defined as follows:

$$P\left(\theta_A \leq \theta_B\mid\textbf{reported results}\right)= P\left(\mu_A \leq \mu_B\mid \widehat{\mu}_A, \widehat{\mu}_B\right)=t_{n-1}\left(  \sqrt{n} \frac{\widehat{\mu}_B - \widehat{\mu}_A}{\sqrt{s_A^2 + s_B^2 - 2 s_A s_B r_{AB}}} \right) $$
where:
\begin{itemize}
    \item $\mu_A$ (resp $\mu_B$) is the mean DSC of method A (resp. method B). 
    \item $n$ is the test set sample size.
    \item $\widehat{\mu}_A$ (resp. $\widehat{\mu}_B$) is the reported mean DSC of method A (resp. method B)
    \item $t_{n-1}$ is the quantile of the Student distribution with $n-1$ degrees of freedom.
    \item $s_A$ (resp. $s_B$) is the standard deviation for method A (resp. for method B) imputed from $\widehat{\mu}_A$ (resp. $\widehat{\mu}_B$).
    \item $r_{AB}$ is the model congruence, i.e., the correlation between the performance of method A and method B. Again, as we do not have access to this value, we make an assumption based on previous experiments. 
\end{itemize}

When applying our formulae to the data collected from the systematic analysis of MICCAI 2023 classification and segmentation papers, certain key parameters required for the calculations were not directly available. To address this, we implemented several strategies to compensate for the missing values, particularly focusing on model congruence.

To get an empirical estimate for model congruence for classification tasks, we selected a diverse set of 12 medical imaging tasks from \cite{godau2024fingerprinting,godau2024mml} covering eight different modalities, namely fundus photography , confocal laser endomicroscopy \cite{aidae2024challenge} , computed tomography \cite{yang2020covidct}, dermatoscopy \cite{kawahara2018seven,isic2020challenge,rotemberg2021melanoma}, gastroscopy and colonoscopy \cite{borgli2020hyperkvasir}, sonography \cite{aldhabyani2020breast}, and X-ray \cite{rajpurkar2017mura,irvin2019chexpert}. Our task selection aimed to cover sufficient variety in modality, number of samples and class imbalance. The  median test set size was 304 images. We then applied a total of 5 \cite{yu2023metaformer,wu2022tinyvit,liu2021swin,dai2021coatnet,yalniz2019billion} different architectures to all tasks and calculated model congruence values for Accuracy from which we retained the median (0.67), 25th percentile (0.47), and 75th percentile (0.83). Since model congruence values for Accuracy are constrained by lower and upper bounds (see Supplementary Information, Subsection C) we adapted the extrapolation of these derived congruence values to the eligible MICCAI 2023 classification papers. The procedure involved first assigning an assumed model congruence value (either the median, 25th percentile, or 75th percentile derived from our private dataset) to each paper. We then assessed whether these extrapolated values fell within the acceptable bounds defined by the characteristics of each MICCAI 2023 classification paper. If an extrapolated value exceeded those bounds, it was adjusted by replacing it with either the lower  or upper bound, selecting whichever value had the smallest difference from the original extrapolated value.

For segmentation papers, we calculated correlation coefficients of the DSC values between all 19 participating methods using the MSD dataset, which comprises 10 tasks over 10 different organs. In our analysis, we used the median correlation coefficient (0.67) from this dataset as the assumed model congruence value for segmentation tasks. To examine the robustness of our findings, we conducted a sensitivity analysis by repeating the probability of false claims calculations with either the 25th percentile (0.44) or 75th percentile (0.82) congruence values from the MSD dataset.

Additionally, to apply the formula for estimating the probability of false claims in segmentation papers, standard deviation (SD) values of the DSC for each of the two methods being compared were required. While the test set sample sizes (N) were generally accessible in the reported data, SD values were frequently missing (see Figure S1). To address this, we imputed the missing standard deviation values following the approach described in \cite{10.1007/978-3-031-72117-5_12}. Similarly to what was done for congruence, we assessed the  sensitivity of the probability of false claims to this imputation. To that purpose, we computed the prediction interval (PI) for the SD imputation and repeated our analysis with the first quartile (Q1) and third quartile (Q3) of the PI  \cite{10.1007/978-3-031-72117-5_12,wasef2016derivation}. Results show  that our conclusions are robust with respect to the imputation of SD (Figure S4).

\section*{Data availability}
The dataset used in this study is not publicly available due to the sensitive nature of the data, which includes information on scientific publications that could affect the reputation of individual authors. To protect confidentiality, the data can be made available from the corresponding author upon reasonable request and subject to approval ensuring responsible use.

\section*{Code availability}
Our code is available at \url{https://github.com/IMSY-DKFZ/probability-of-false-claims}.


\end{bibunit}

\section*{Acknowledgments}

This publication received funding from the European Research Council (ERC) under the European Union’s Horizon 2020 research and innovation program (grant agreement no. 101002198, NEURAL SPICING). Part of this work was also funded by Helmholtz Imaging (HI), a platform of the Helmholtz Incubator on Information and Data Science.
Moreover, this project has received funding from the National Center for Tumor Diseases (NCT) Heidelberg’s Surgical Oncology Program.
This publication was also supported through state funds approved by the State Parliament of Baden-Württemberg for the Innovation Campus Health + Life Science Alliance Heidelberg Mannheim.
The research leading to these results has received further funding from the French government under management of Agence Nationale de la Recherche as part of the “France 2030” program (reference ANR-23-IACL-0008, project PRAIRIE-PSAI) and as part of the "Investissements d'avenir" program (reference ANR-19-P3IA-0001, project PRAIRIE 3IA Institute, reference ANR-10-IAIHU-06, project Agence Nationale de la Recherche-10-IA Institut Hospitalo-Universitaire-6), from the European Union’s Horizon Europe Framework Programme (grant number 101136607, project CLARA), from Agence Nationale de la Recherche (reference ANR-23-CE45-0005-01, project ANO-NEURO) and from Inserm and the French Ministry of Health in the context of the MESSIDORE 2023 call operated by IReSP (reference AAP-2023-MSDR-341011).
The research has also been funded by the Ministry of Education, Youth and Sports of the Czech Republic (Project LM2023050).

\section*{Author Contributions}

\begin{itemize}
    \item Initialization of study: OC and LMH
    \item Conceptualization: EC, OC, LMH, AR, GV
    \item Probability of false claims derivations: PA, OC, AG, CH
    \item Article screening: PA, SB, NB, VC, OC, SE, RH, PK, MK , SL, NR, CS, MS
    \item Conflict resolution: EC, AR
    \item Development of data analysis tools: EC, OC
    \item Data analysis: EC, OC
    \item Writing of paper: EC, OC, LMH, AR, GV
    \item Revision of paper: all
    \item Figures: EC, AR 
\end{itemize}

\section*{Competing Interests}

Competing financial interests unrelated to the present article: OC reports having received consulting fees from Therapanacea. OC reports that other principal investigators affiliated to the team which he co-leads have received grants (paid to the institution) from Sanofi and Biogen. OC reports that his spouse was an employee of myBrainTechnologies (2015-2023) and is an employee of DiamPark. 

NR is an employee of NVIDIA.


\begin{bibunit}[naturemag]

\section*{Supplementary Information}



\renewcommand{\thetable}{S\arabic{table}}
\renewcommand{\thefigure}{S\arabic{figure}}

\appendix

\section{Results about practices in the community}\label{suppl:A}

\begin{figure}[H]
\centering
\includegraphics[width=0.8\textwidth]{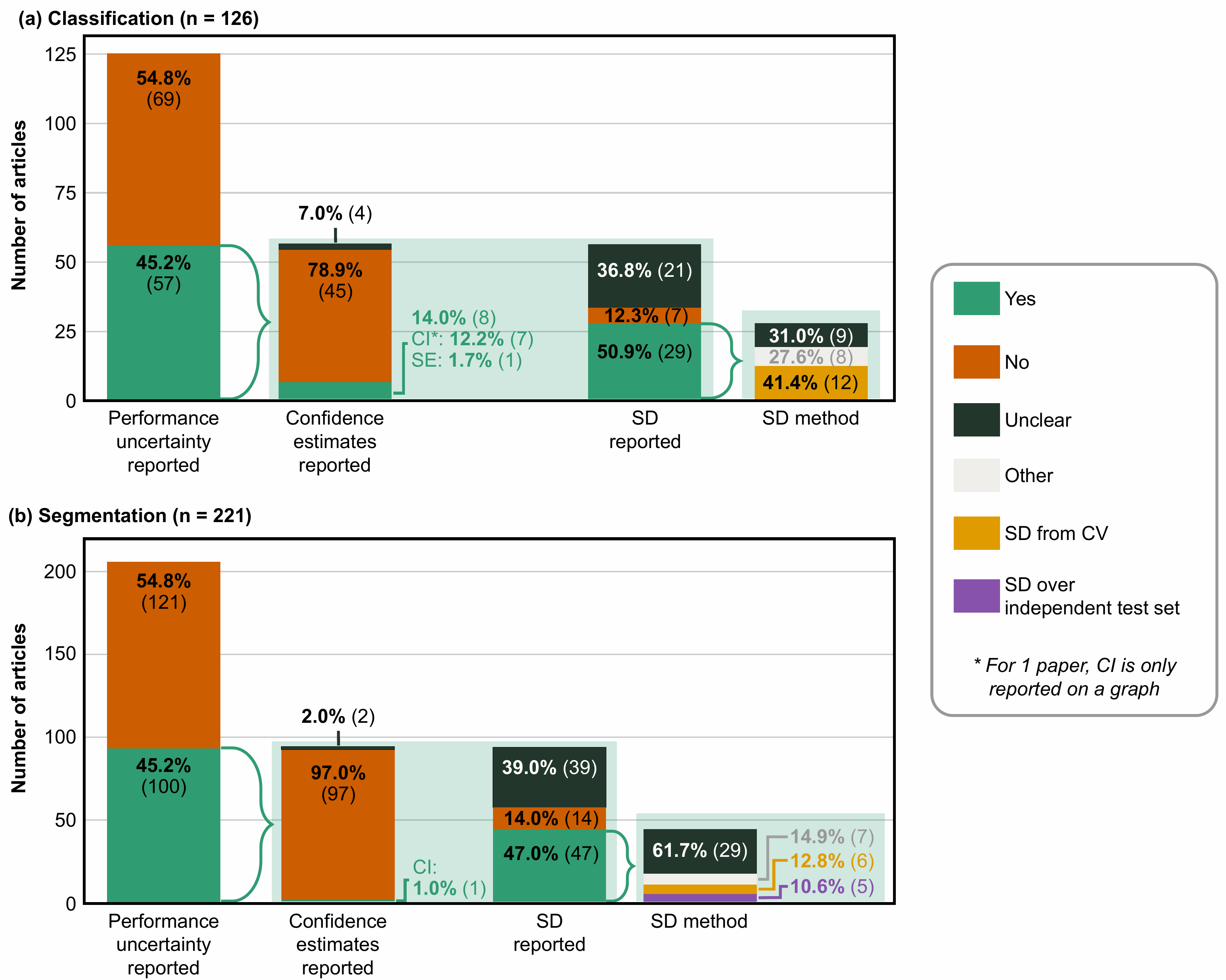} 
\caption{\textbf{\textit{The majority of MICCAI 2023 papers ($>$50\%) are not reporting performance uncertainty.}} Barplots demonstrating practices of performance uncertainty reporting in MICCAI 2023 classification (a) and segmentation (b) papers. 
\\SD; standard deviation, CV; cross-validation, CI; confidence interval, MICCAI; Medical Image Computing and Computer-Assisted Intervention conference}
\label{FigS1}
\end{figure}

Figure \ref{FigS1} summarizes the reporting practices for performance uncertainty. Over half of the papers did not report any uncertainty estimates, such as standard deviation (SD) or confidence intervals (CIs). Among those that did, SD was the most common, reported by 51\% of classification papers and 47\% of segmentation papers. However, only five segmentation papers—and none in classification—reported SD on an independent test set. In other cases, SD was either computed across cross-validation (CV) folds (a biased estimator \cite{bengio2004no}), over different training configurations, or using unclear methods, making it unsuitable for inferential statistics. Notably, confidence intervals were reported in just 12\% of classification papers and 1\% of segmentation papers that reported performance uncertainty.

\begin{figure}[H]
\centering
\includegraphics[width=0.8\textwidth]{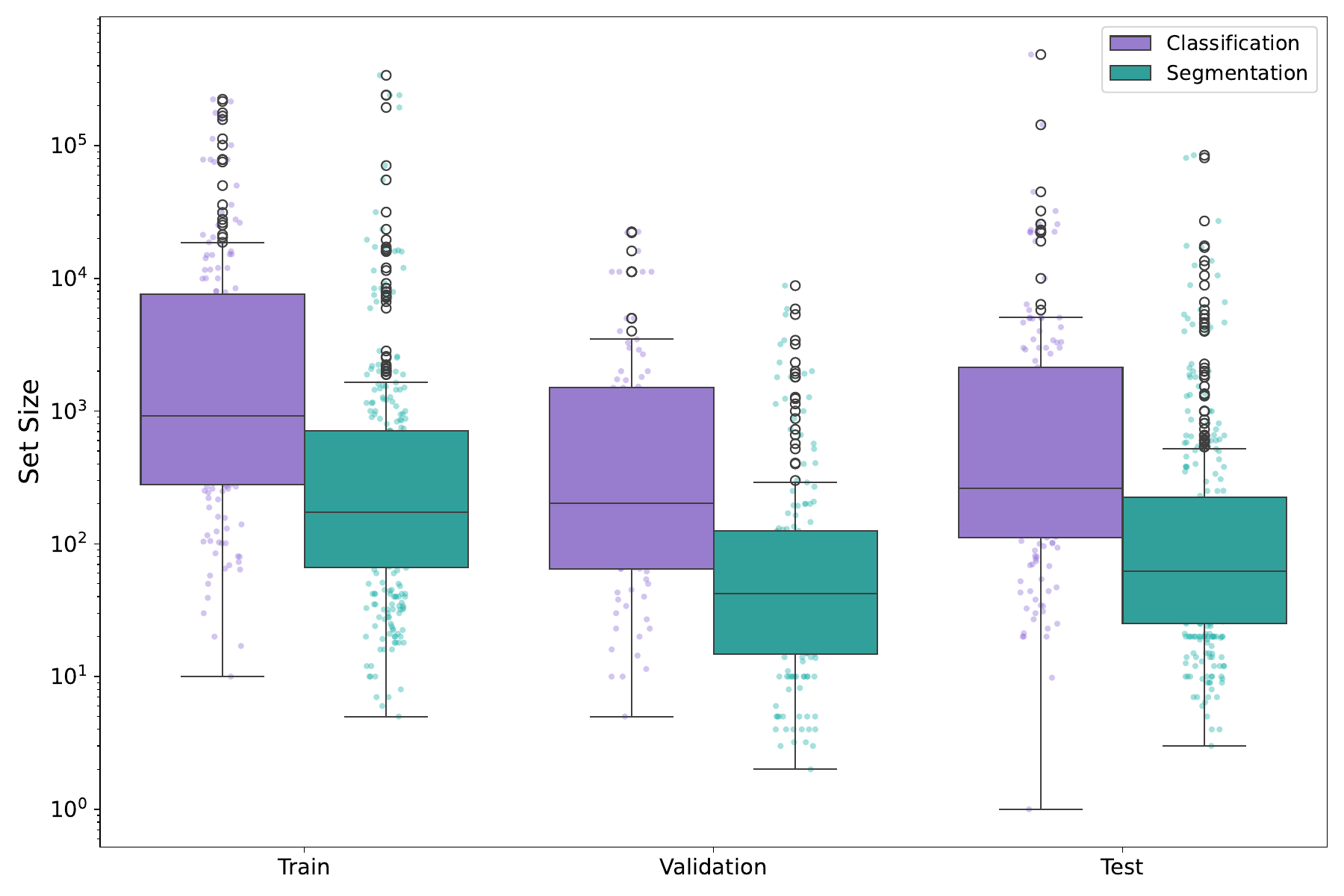} 
\caption{\textbf{\textit{Number of train, validation, and test images for MICCAI 2023 classification and segmentation papers.}} For experiments conducted within a paper that were using k-fold CV as a data splitting approach, the size of the train and set sizes were defined as: (k-1)*N/k and N/k respectively. Box bounds represent the 25th (Q1) and 75th (Q3) percentiles; whiskers extend to max(min, $Q1–1.5×IQR)$ and min(max, $Q3+1.5×IQR$). 
\\k; number of folds, CV; cross-validation, N; number of images of the full set; Q1; 25th percentile, Q3; 75th percentile, IQR; interquartile range, MICCAI; Medical Image Computing and Computer-Assisted Intervention conference}
\label{FigS2}
\end{figure}

\vspace{1cm}

\begin{table}[H]
\centering
\caption{\textbf{\textit{Accuracy is the most frequently reported metric for classification papers of MICCAI 2023.}}}
\begin{tabular}{l c} \hline
\textbf{Metric} & \textbf{Percentage (\%)} \\ \hline
Accuracy & 72.4 \\
Area Under the Curve (AUC) & 54.5 \\
F1 score & 41.5 \\
Sensitivity & 35.0 \\
Positive Predictive Value (PPV) & 19.5 \\
Specificity & 16.3 \\
Balanced Accuracy (BA) & 8.9 \\
Expected Calibration Error (ECE) & 3.3 \\
(Weighted) Cohen's Kappa ((W)CK) & 2.4 \\
Average Precision (AP) & 1.6 \\
Mean Absolute Error (MAE) & 0.8 \\
Negative Predictive Value (NPV) & 0.8 \\ \hline
\end{tabular}
\label{tabS1:classification_metrics}
\end{table}

\vspace{1cm}

\begin{table}[H]
\centering
\caption{\textbf{\textit{Dice Similarity Coefficient (DSC) is the most frequently reported metric for segmentation papers of MICCAI 2023.}}}
\begin{tabular}{l c} \hline
\textbf{Metric} & \textbf{Percentage (\%)} \\ \hline
Dice Similarity Coefficient (DSC) & 94.8 \\
Intersection over Union (IoU)/Jaccard Index & 35.8 \\
Hausdorff Distance (HD) & 29.2 \\
Average Surface Distance (ASD) & 10.4 \\
Hausdorff Distance 95 (HD95)\% & 3.3 \\
Normalized Surface Distance (NSD) & 1.9 \\
Maximum Symmetric Surface Distance (MSSD) & 0.5 \\
Fbeta score & 0.5 \\ \hline
\end{tabular}
\label{tabS2:segmentation_metrics}
\end{table}

\vspace{1cm}

\section{Variability in model performance differences of MICCAI 2023 papers}\label{suppl:B}

\begin{figure}[H]
\centering
\includegraphics[width=0.8\textwidth]{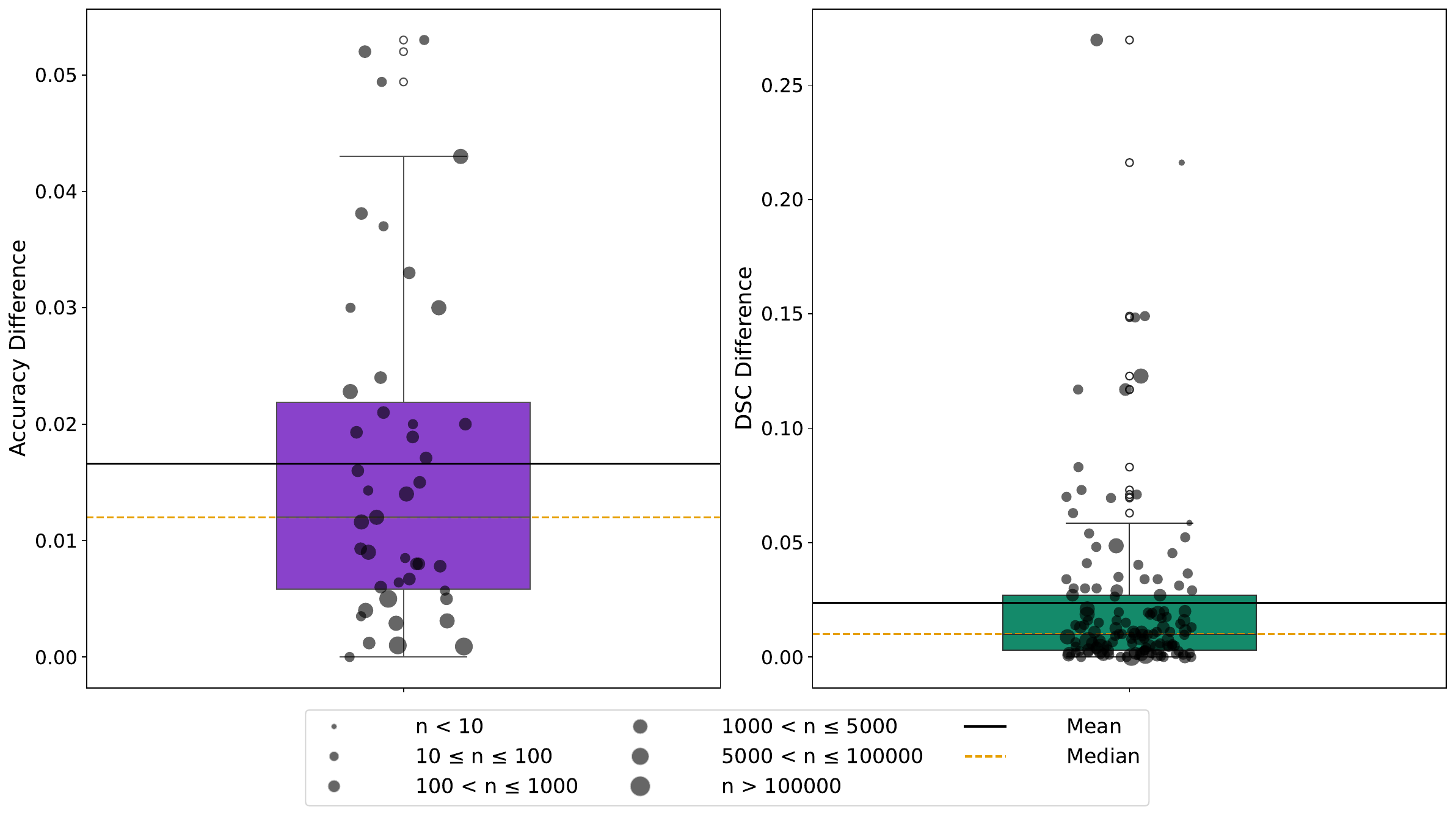} 
\caption{\textbf{\textit{Performance difference between the two top-ranked methods of eligible MICCAI 2023 classification and segmentation papers.}} Dot sizes indicate test set size categories. Box bounds represent the 25th (Q1) and 75th (Q3) percentiles; whiskers extend to max(min, $Q1–1.5×IQR)$ and min(max, $Q3+1.5×IQR$). The solid black line shows the mean performance difference (Accuracy: 0.02, DSC: 0.02), and the dashed orange line shows the median (Accuracy: 0.01, DSC: 0.01).
\\k; number of folds, CV; cross-validation, n; number of images of the full set; Q1; 25th percentile, Q3; 75th percentile, IQR; interquartile range, DSC; Dice Similarity Coefficient, MICCAI; Medical Image Computing and Computer-Assisted Intervention conference}
\label{FigS3}
\end{figure}

\section{Probability of false claims formulas}\label{suppl:C}

\subsection{Problem statement}

Here, we provide a detailed derivation of the ``probability of false claims", i.e., the probability that the second-ranked method was in fact as good or better than the first-ranked method, given the results reported in the paper. As described in the main text, this is defined as:
$$
P(\theta_A \leq \theta_B | \textbf{reported results})=P(\theta_A \leq \theta_B | \widehat{\theta}_A, \widehat{\theta}_B)$$

where $\theta_A$ (resp. $\theta_B$) is the true performance of the first-ranked method A (resp. second-ranked method B) and $\widehat{\theta}_A$ (resp. $\widehat{\theta}_B$) is the performance of method A (resp. method B) reported in the paper (in other words, the observed performance). 
Following a classical Bayesian approach, $\theta_A$ and $\theta_B$ are treated as random variables while the observed values $\widehat{\theta}_A$ and $\widehat{\theta}_B$ are treated as fixed. Note that, by definition, the probability of false claims cannot be greater than $0.5$, since one has observed that $\widehat{\theta}_A>\widehat{\theta}_B$.

To obtain this, we first derive $P(\theta_A \leq \theta_B | \bm{x})$ where $\bm{x}$ are the results on the test set. Of course, we do not have access to these, but only to the information provided in the paper. We will then show that $P(\theta_A \leq \theta_B | \bm{x})$ can be expressed using only the information provided in the paper and assumptions on the model congruence.

In the following, we provide two derivations of the probability of false claims: 
\begin{itemize}
    \item for classification tasks: the performance is the accuracy, i.e., the probability that an observation is well classified, and we classically write $\theta= p$.
    \item for segmentation tasks: the performance is the mean Dice Similarity Coefficient (DSC) and we classically write $\theta=\mu$.
\end{itemize} 

\subsection{Formula for classification}

\subsubsection{Definitions}
Let $n$ be the test set size and $x_A$ (resp. $x_B$) a random variable representing, for a given  observation, whether it is correctly classified by method A (resp. method B). $x_A$ (resp. $x_B$) follows a Bernoulli distribution with parameter $p_A$ (resp. $p_B$) where $p_A=P\left(x_A=1\right)$ is the probability that an observation is well
classified by method A, in other words, the true accuracy of method A (resp. $p_B= P\left(x_B=1\right)$). The following contingency table describes the agreement between $x_A$ and $x_B$\footnote{Note that this is not the contingency table of one classifier with respect to the ground-truth.}:
\begin{center}

        \begin{tabular}{c|c|c|c}
            \multicolumn{1}{c}{}&\multicolumn{1}{c}{$x_A=0$}&\multicolumn{1}{c}{$x_A=1$}\\
            \cline{2-3}
             $x_B=0$& $x_{0,0}$ & $x_{1,0}$ & $n-n_B$ \\
            \cline{2-3}
             $x_B=1$& $x_{0,1}$ & $x_{1,1}$ & $n_B$ \\
            \cline{2-3}
             \multicolumn{1}{c}{}&\multicolumn{1}{c}{$n-n_A$} &\multicolumn{1}{c}{$n_A$} &$n$\\
           
        \end{tabular}
\end{center}

Similarly, one can define the following probabilities:
\begin{align*}
    p_{1,1}&= P\left(x_A=1, x_B=1\right)\\
    p_{1,0}&= P\left(x_A=1, x_B=0\right)\\
    p_{0,1}&= P\left(x_A=0, x_B=1\right)\\
    p_{0,0}&= P\left(x_A=0, x_B=0\right)
\end{align*}

\noindent Let 
 $\bm{x}= \left(x_1, x_2, x_3, x_4\right)=\left(x_{1,0},x_{0,1},x_{1,1}, x_{0,0}\right) $ 
and $\left(p_1, p_2, p_3, p_4\right)=\left(p_{1,0},p_{0,1},p_{1,1}, p_{0,0}\right)$

\noindent One can note that
\begin{itemize}
    \item $p_A= p_{1,1} + p_{1,0}=p_1+p_{1,1}$
    \item $p_B= p_{1,1} + p_{0,1}=p_2+p_{1,1}$
\end{itemize}

\noindent Thus: 
 $P(p_A\leq p_B |\bm{x})=P(p_1 \leq p_2|\bm{x})$.
To compute this probability, we need the posterior probability density function (pdf) $p(p_1, p_2| \bm{x})$.  
\vspace{0.5cm}
 
\noindent Note that, by construction, we have two constraints on $p_{1,1}$ which come from the contingency table:
\begin{itemize}
\item $ p_A + p_B - 1  \leq p_{1,1}$ (i.e., two classifiers with a certain accuracy on the test sample must share at least a certain proportion of their accurate predictions)
\item $p_{1,1} \leq \min(p_A,p_B)$ (i.e., a classifier cannot have a proportion of accurate samples shared with the other classifier greater than its own accuracy) 
\end{itemize}

In summary:
$ p_A + p_B - 1  \leq p_{1,1} \leq \min(p_A,p_B)$

\subsubsection{Likelihood} 

Since $p_{1,1} + p_{1,0} + p_{0,1} + p_{0,0}= p_1 + p_2+ p_3 + {p_4} =1$ and $x_{1,1}+ x_{1,0} +x_{0,1}+ x_{0,0}= x_1 + x_2 + x_3 +{x_4} =n$ then $\bm{x}$ follows a multinomial distribution, whose probability density function (pdf) is denoted as $M$, with parameters $p_1, p_2, p_3,{p_4}$. Thus, the likelihood is:
\begin{align*}
    p(\bm{x}|p_1, p_2, p_3,{p_4})
    &=M(p_1, p_2, p_3, {p_4})\\
    &= \frac{n!}{x_1! x_2!x_3!{x_4!}} p_1^{x_1}p_2^{x_2}p_3^{x_3}{p_4^{x_4}}
\end{align*}

\subsubsection{Prior}
 Classically, we choose a prior Dirichlet distribution since it is the conjugate prior of the multinomial distribution (\cite{lesaffre2012bayesian}, section 5.3.2): 

\begin{align*}
    p(p_1, p_2, p_3,{p_4})&=D(\alpha_1 ,\alpha_2, \alpha_3, {\alpha_4})\\
    &= \frac{1}{\mathcal{B}\left(\alpha_1, \alpha_2, \alpha_3,{\alpha_4}\right)} p_1^{\alpha_1-1} p_2^{\alpha_2-1}p_3^{\alpha_3-1}{p_4^{\alpha_4-1}}
\end{align*}
We further choose an uninformative prior by setting $ \alpha_1 = \alpha_2= \alpha_3= {\alpha_4} =1$ (\cite{lesaffre2012bayesian}, Section 4.4.2, bottom of page 91), resulting in:
$$p(p_1, p_2, p_3,{p_4})=  \frac{1}{\mathcal{B}\left(1,1,1,{1}\right)}$$

\subsubsection{Posterior}
From Bayes' formula:
\begin{align*}
    p(p_1, p_2, p_3,{p_4}|\bm{x})& \propto p(\bm{x}|p_1, p_2, p_3, {p_4})p(p_1, p_2, p_3, {p_4})\\
    &\propto \frac{n!}{x_1! x_2!x_3!{x_4!}} p_1^{x_1}p_2^{x_2}p_3^{x_3}{p_4^{x_4}} \times \frac{1}{\mathcal{B}\left(1,1,1,{1}\right)}\\
    & \propto D\left(x_1 +1, x_2 +1, x_3 +1, {x_4 + 1}\right)
\end{align*}

The probability of false claims is:
 \begin{align*}
{P(p_1 \leq p_2 \mid \bm{x})} & = \int_{0}^{1} \int_{0}^{p_2}p(p_1, p_2|\bm{x}) dp_1 dp_2 \\
\end{align*}

Thus, we now need $p(p_1, p_2|\bm{x})$.  

 First, we use the fact that {$p_4= 1- p_1 - p_2 - p_3$} and {$x_4= n- x_1 - x_2 - x_3$}, resulting in:
\begin{align*}
    {p(p_1, p_2, p_3|\bm{x})} & \propto {D\left(x_1 +1, x_2 +1, x_3 + 1,  n- x_1 - x_2 - x_3 +1\right)}
\end{align*}

Then we have to marginalize $p(p_1, p_2, p_3|\bm{x})$ over $p_3$ by integrating: 

   $$ p(p_1, p_2|\bm{x}) = 
\int_{0}^{1} p(p_1, p_2,p_3|\bm{x}) dp_3$$

Using the fact that $p_3 = 1 - p_1 - p_2 - p_4$, $p_3$ does not exist in the $[1-p_1-p_2-p_4, 1]$ interval. Thus, in particular, $p_3$ does not exist in the $[1-p_1-p_2, 1]$ interval. Therefore:

\begin{align*}
    p(p_1, p_2|\bm{x}) &= 
\int_{0}^{1-p_1-p_2} p(p_1, p_2,p_3|\bm{x}) dp_3\\
&= \frac{1}{B(x_1 +1, x_2 +1, x_3 +1, n - x_1 - x_2 - x_3 +1)}   p_1^{x_1} p_2^{x_2} \int_{0}^{1-p_1-p_2} p_3^{x_3} (1-p_1-p_2-p_3)^{n-x_1 - x_2 - x_3} dp_3 
\end{align*}
 
Let $a = (1-p_1-p_2)$ and use the change of variable $u = \frac{p_3}{a}, \ dp_3=adu$: 

 \begin{align*}
       \int_{0}^{1-p_1-p_2} p_3^{x_3} (1-p_1-p_2-p_3)^{n-x_1 - x_2 - x_3} dp_3& = \int_{0}^{a} p_3^{x_3} (a-p_3)^{n-x_1 - x_2 - x_3} dp_3 \\
       &  = \int_{0}^{1} (ua)^{x_3} (a-ua)^{n-x_1 - x_2 - x_3}adu \\
       & = a^{n-x_1-x_2+1} \int_{0}^{1} u^{x_3} (1-u)^{n-x_1 - x_2 - x_3}du \\
       & = (1-p_1-p_2)^{n-x_1-x_2+1}B(x_3+1, n-x_1-x_2-x_3+1)
 \end{align*}

\noindent This leads to :
$$
 p(p_1, p_2|\bm{x}) = \frac{B(x_3, n-x_1-x_2-x_3)}{B(x_1 +1, x_2 +1, x_3 +1, n - x_1 - x_2 - x_3 +1)}p_1^{x_1}p_2^{x_2}(1-p_1-p_2)^{n-x_1-x_2+1}
$$

By definition:
\[
B(x_3+1, n-x_1-x_2-x_3+1) =
\frac{\Gamma(x_3+1) \Gamma(n-x_1-x_2-x_3+1)}{\Gamma(n-x_1-x_2+2)},
\]
and
\[
\frac{1}{B(x_1+1, x_2+1, x_3+1, n-x_1-x_2-x_3+1)} =
\frac{\Gamma(n+4)}{\Gamma(x_1+1) \Gamma(x_2+1) \Gamma(x_3+1) \Gamma(n-x_1-x_2-x_3+1)}.
\]

\noindent As a result, one has:
\begin{align*}
{p(p_1, p_2|\bm{x})} &= \frac{\Gamma(n+4)}{\Gamma(x_1+1)\Gamma(x_2+1)\Gamma(n-x_1-x_2+2)} 
p_1^{x_1} p_2^{x_2} (1-p_1-p_2)^{n-x_1-x_2+1}, \\
 &= \frac{1}{B(x_1+1,x_2+1,n-x_1-x_2+2)}p_1^{x_1} p_2^{x_2} (1-p_1-p_2)^{n-x_1-x_2+1} \\
 &= {D\left(x_1 + 1, x_2 + 1, n - x_1 - x_2 + 2\right)}.
\end{align*}

\subsubsection{Probability of false claims}

As one can see, the integral $\int_{0}^{1} \int_{0}^{p_2}p(p_1, p_2|\bm{x}) dp_1 dp_2$ is not tractable analytically.  It was therefore computed using Monte Carlo sampling from the Dirichlet distribution. We can express the parameters $x_1$ and $x_2$ of the Dirichlet as follows.

For a given paper, one has:
\begin{itemize}
\item $\widehat{p}_A= \frac{n_A}{n}$ the reported accuracy of method A 
\item $\widehat{p}_B= \frac{n_B}{n}$ the reported accuracy of method B
\end{itemize}

Further, we make an assumption on the model congruence $\widehat{p}_{1,1}$, i.e., is the proportion of cases where both methods made correct predictions (please refer to Methods section for details).

\noindent This gives:
\begin{itemize}
\item $x_1=n(\widehat{p}_A-\widehat{p}_{1,1})$
\item $x_2=n(\widehat{p}_B-\widehat{p}_{1,1})$
\end{itemize}

We then sample $k$ times $p_1,p_2$ from {$\mathcal{D}\left(x_1 +1, x_2 +1,  n-x_1 - x_2 +2\right)$}, and count the number of times $M$ where $p_1\leq p_2$.
 
We thus obtain the probability of false claims $$P(\theta_A\leq \theta_B |\widehat{\theta}_A, \widehat{\theta}_B)=P(p_A\leq p_B |\widehat{p}_A,\widehat{p}_B)\approx \frac{M}{k}$$

\subsection{Formula for segmentation}
  \subsubsection{Definitions}
  Let's define the following:
 \begin{itemize}
     
     \item $n$ be the test set size and .
    \item $x_A$ (resp. $x_B$) a random variable representing, for an  observation $y$ of the test set, the performance of method A (resp. method B), as measured by the DSC with respect to the reference. 
     \item $x = x_A - x_B$ the difference in DSC between methods A and B.
         \item $\bm{x}=(x_1,\ldots,x_n)$ a random sample of size $n$ of $x$, from which we can compute $\bar{x}= \frac{1}{n}\sum_{i=1}^n x_i$.
 \end{itemize}
 We assume that $x_A$ and $x_B$ are jointly normal~\footnote{This assumption is necessary so that the difference is normally distributed, since $x_A$ and $x_B$ are not independent.}, i.e., that $(x_A,x_B) \sim \mathcal{N}(\bm{M},\bm{\Sigma})$ with $\bm{M}=\begin{pmatrix} \mu_A \\ \mu_B \end{pmatrix}$ 
and $\bm{\Sigma}=\begin{pmatrix} \sigma_A^2 & \sigma_A \sigma_B \rho_{AB} \\ \sigma_A \sigma_B \rho_{AB} & \sigma_B^2 \end{pmatrix}$.
  We then have $x\sim \mathcal{N}(\mu, \sigma^2)$ 
with $\mu = \mu_A - \mu_B$  and $ \sigma^2 = \sigma_A^2 + \sigma_B^2 - 2 \sigma_A \sigma_B \rho_{AB} $.

Then
$$
P(\mu_A \leq\mu_B | \bm{x})=P(\mu \leq 0 | \bm{x})$$

To compute this probability, we need the posterior pdf $p(\mu | \bm{x})$.  

The following derivations are classical Bayesian statistics and can for instance be found in \cite{gelman1995bayesian}, (Section 3.2, page 79) and 
\cite{lesaffre2012bayesian}. (Section 4.3.1, page 85). We recall them here for the sake of completeness.

\subsubsection{Likelihood}

Let $N$ be the pdf of the normal distribution $\mathcal{N}(\mu,\sigma^{2})$, one has:
 \begin{align*}
     p(\bm{x} \mid \mu, \sigma^{2})&=\prod_{i=1}^{n}N\left(x_{i} ; \mu, \sigma^{2}\right) \\
     &= \frac{1}{\sqrt{2\pi}^n}\sigma^{-n} \exp\left(-\frac{1}{2\sigma^{2}}\left(\sum_{i=1}^n x_i^2 - 2 \sum_{i=1}^nx_i \mu + \sum_{i=1}^n \mu^2\right)\right)\\
\end{align*}

Let:
\begin{itemize}
    \item $\bar{x}= \frac{1}{n}\sum_{i=1}^n x_i$ 
    \item $s^2=\frac{1}{n-1}\sum_{i=1}^n (x_i -\bar{x})^2=
    \frac{1}{n-1}\sum_{i=1}^n (x_i^2 -2x_i\bar{x}+\bar{x}^2)=
    \frac{1}{n-1}(\sum_{i=1}^n x_i^2 - 2\bar{x}\sum_{i=1}^n x_i +\sum_{i=1}^n \bar{x}^2)=
    \frac{1}{n-1}(\sum_{i=1}^n x_i^2 - 2n\bar{x}^2 + n\bar{x}^2)=
    \frac{1}{n-1}\sum_{i=1}^n (x_i^2 -2\bar{x}^2+\bar{x}^2)=
    \frac{1}{n-1}\sum_{i=1}^n (x_i^2 -\bar{x}^2)$.
\end{itemize}

Then:
\begin{align*}
p(\bm{x} \mid \mu, \sigma^{2}) 
     &=\frac{1}{\sqrt{2\pi}^n}\sigma^{-n} \exp\left(-\frac{1}{2\sigma^2}\left((n-1)s^2+n(\mu - \bar{x})^2\right)\right)
 \end{align*}


\subsubsection{Prior}

We choose the following classical non-informative joint prior: $p(\mu,\sigma^2)\propto \sigma^{-2}$ (see e.g. \cite{gelman1995bayesian}, Section 3.2, page 76).

\subsubsection{Posterior}
According to Bayes formula, we can compute the posterior distribution of $\mu, \sigma^2 | \bm{x}$.
\begin{align*}
    p(\mu, \sigma^{2} |\bm{x})
    & \propto\sigma^{-n-2} \exp\left(- \frac{1}{2\sigma^{2}}\left( {(n-1)s^2} +{n(\bar{x}-\mu)^2}\right)\right)
\end{align*}

We now have to integrate $p(\mu, \sigma^{2} |\bm{x})$ over $\sigma^{2}$ to obtain $p(\mu | \bm{x})$. \\
\begin{align}
    p(\mu | \bm{x})&= \int_{\mathbb{R}^+} p(\mu, \sigma^{2} |\bm{x})d\sigma^{2}\\
    &\propto \int_{\mathbb{R}^+} \sigma^{-n-2} \exp\left(- \frac{1}{2\sigma^2}\left( {(n-1)s^2} +{n(\bar{x}-\mu)^2}\right)\right)d\sigma^{2}\label{eqn:post1}
\end{align}
This integral can be evaluated using the following  substitution (provided by \cite{gelman1995bayesian}, section 3.2, page 78)
\[
z = \frac{A}{2\sigma^2}, \quad \text{where} \quad A = (n-1)s^2 + n(\mu - \bar{x})^2
\]

Thus equation~\ref{eqn:post1} becomes:

\begin{align*}
 p(\mu | \bm{x} )   &   \propto A^{-n/2} \int_0^\infty z^{(n/2-1)} \exp(-z) \, dz\\
 & \propto T\left(\frac{\mu -\bar{x}}{\sqrt{\frac{s^2}{n}} }; n-1\right)
\end{align*}

where $T$ is the pdf of the Student distribution.

Using the fact that if $Z$ is a variable of density $g$ then the density of $\frac{Z-a}{b}$ is $u\mapsto bg(bu +a)$, we have: 

$$ p\left(\frac{\mu -\bar{x}}{\sqrt{\frac{s^2}{n}}} | \bm{x} \right)  \propto T\left(\mu ; n-1\right)$$

\subsubsection{Probability of false claims}

$$ P\left(\mu \leq 0 \mid \boldsymbol{x}\right)= P\left(\frac{\mu -\bar{x}}{\sqrt{\frac{s^2}{n}} } \leq \frac{-\bar{x}}{\sqrt{\frac{s^2}{n}} } \mid \boldsymbol{x}\right)=t_{n-1}\left(\frac{-\bar{x} \sqrt{n}}{\sqrt{\frac{\sum_{i=1}^{n}\left(x_i-\bar{x}\right)^{2}}{n-1}}} \right) $$

where $t_{n-1}(z)$ denotes the cumulative distribution of the standardized Student distribution with $n-1$ degrees of freedom computed at $z$.

One has:
\begin{itemize}
    \item $\bar{x}=\widehat{\mu}_A-\widehat{\mu}_B$ where $\widehat{\mu}_A$ and $\widehat{\mu}_B$ are the observed mean performances reported in the paper.
    \item $\frac{\sum_{i=1}^{n}\left(x_i-\bar{x}\right)^{2}}{n-1}\approx s_A^2 + s_B^2 - 2 s_A s_B r_{AB}$ where $s_A$ and $s_B$ are the imputed standard deviation (SD) values, and $ r_{AB}$ is the assumed correlation coefficient value, which in our main text is described as model congruence (please refer to Methods section for details on SD imputation and congruence).
\end{itemize}

We then have the probability of false claims:
$$ P(\theta_A \leq \theta_B | \widehat{\theta}_A, \widehat{\theta}_B)=P\left(\mu_A \leq \mu_B \mid \widehat{\mu}_A, \widehat{\mu}_B\right)
= t_{n-1}\left(  \sqrt{n} \frac{\widehat{\mu}_B-\widehat{\mu}_A}{\sqrt{s_A^2 + s_B^2 - 2 s_A s_B r_{AB}}} \right)$$

\section{Sensitivity analysis for imputation of SD}\label{suppl:D}

\begin{figure}[H]
\centering
\includegraphics[width=0.8\textwidth]{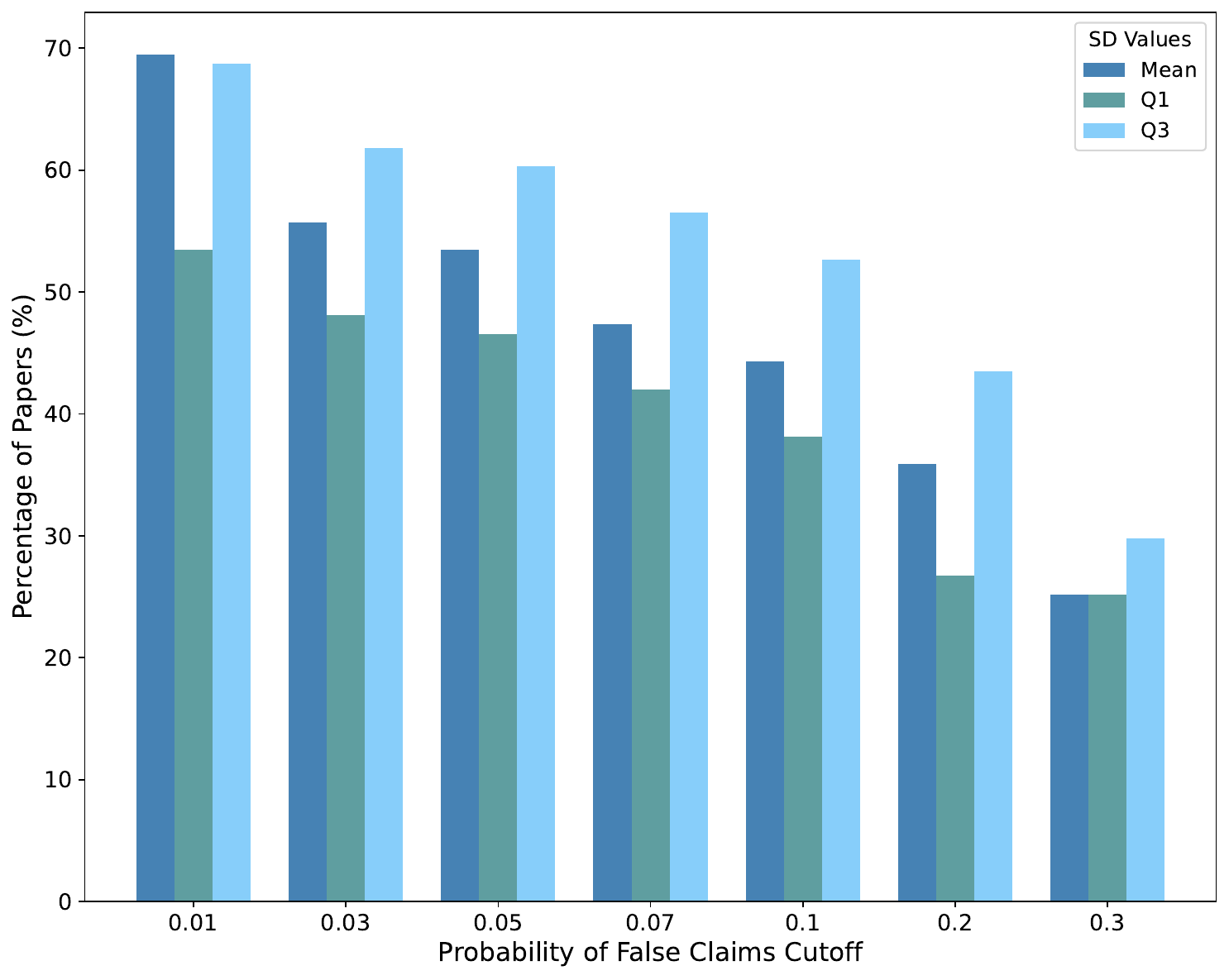} 
\caption{\textbf{\textit{The high likelihood for false outperformance claims in medical image segmentation is robust with respect to various values of imputed standard deviation and probability of false claims cutoffs.}} Cumulative percentage of MICCAI 2023 segmentation papers (n=131) that are exceeding various cutoffs of the probability of false claims for a variety of  predicted standard deviation values (mean, Q1 or Q3). 
\\SD; standard deviation, Q1; 25th percentile (lower bound of prediction interval), Q3; 75th percentile (upper bound of prediction interval), MICCAI; Medical Image Computing and Computer-Assisted Intervention conference
}
\label{FigS4}
\end{figure}



\renewcommand{\refname}{Supplementary Information References}

\end{bibunit}

\end{document}